%% file: main.tex
\documentclass[pmlr,twocolumn,10pt]{jmlr} 

\usepackage{booktabs}
\usepackage{siunitx}

\usepackage[switch]{lineno}
\usepackage{soul}
\usepackage{enumitem}
\usepackage{algorithm}
\usepackage{algorithm2e}
\usepackage{multirow}
\usepackage{amsmath}

\newcommand{\barwidth}{1.8} 
\newcommand{\barheight}{4.2pt} 
\newcommand{\percentscale}{100} 

\def\bar#1{
   {\color{gray}\rule{\fpeval{#1/\percentscale*\barwidth} cm}{\barheight}} #1
}

\theorembodyfont{\upshape}
\theoremheaderfont{\scshape}
\theorempostheader{:}
\theoremsep{\newline}

\jmlrvolume{333}
\jmlryear{2026}
\jmlrsubmitted{LEAVE UNSET}
\jmlrpublished{LEAVE UNSET}
\jmlrworkshop{Conference on Health, Inference, and Learning (CHIL) 2026} 

\title[SocialLM]{SocialLM: Social Signal Processing of Patient-Provider Communication using LLMs and Contextual Aggregation}

\author{%
\Name{Manas Satish Bedmutha} \Email{mbedmutha@ucsd.edu}\\
\addr UC San Diego, USA
\AND
\Name{Feng Chen} \Email{fengc9@uw.edu}\\
\addr University of Washington, USA
\AND
\Name{Andrea L Hartzler} \Email{andreah@uw.edu}\\
\addr University of Washington, USA
\AND
\Name{Trevor Cohen} \Email{cohenta@uw.edu}\\
\addr University of Washington, USA
\AND
\Name{Nadir Weibel} \Email{weibel@ucsd.edu}\\
\addr UC San Diego, USA
}

\begin{document}

\maketitle

\begin{abstract}

Effective patient–provider communication is difficult to assess at scale. We examine whether large language models (LLMs) can track 20 social behaviors from clinical transcripts without fine-tuning. Across three model families and multiple prompting strategies, LLMs reliably detect social signals, though performance varies by patient race and visit segment. To address this variability under query-only API constraints, we introduce an agreement-weighted ensemble using group-level agreement patterns. This approach improves both accuracy and stability over the best individual model, demonstrating a practical pathway for scalable social signal tracking in clinical conversations.
\end{abstract}

\paragraph*{Data and Code Availability}
We use conversations from the \textit{Establishing Focus} (EF) dataset~\citep{AHRQ2006}, which cannot be publicly shared due to the usage of real-world recordings of patient-provider conversations in clinic visits; interested researchers may contact the authors regarding data-use agreements. Code available [\href{:https://github.com/chenfeng1234567/SocialLM}{here}].

\paragraph*{Institutional Review Board (IRB)}
This secondary analysis is approved by the IRBs of collaborating institutions and is HIPAA-compliant (UW STUDY00005436).

\input{paper/1-introduction}
\input{paper/2-related-work}

\input{paper/3-methods}

\input{paper/4-analyses}

\input{paper/5-ensemble}
\input{paper/6-discussion}
\input{paper/7-conclusion}

\acks{This study was a part of the UnBIASED project which was funded by the National Institutes of Health, National Library of Medicine, project R01LM013301. We thank the funding agency and all of our collaborators on the UnBIASED project, especially co-investigators Wanda Pratt, Janice Sabin and Brian Wood. Bedmutha was also supported by research fellowships from the Jacobs Center for Health Innovation and the Sanford Institute for Empathy and Compassion at UC San Diego.}

\bibliography{sociallm}
\
\appendix
\input{paper/A-Prompt}

\end{document}

%% file: paper/1-introduction.tex
\section{Introduction}
\label{sec:intro}
Communication between patients and primary-care providers (henceforth \textit{providers}) is critical to patient wellbeing.  Good communication improves trust, adherence and visit-satisfaction, thereby improving health outcomes (\cite{street2007physicians}). Clinical conversations are different from everyday conversations due to inherent power asymmetries, time constraints and high stakes, thereby making even subtle interpersonal dynamics consequential. As a result, there is growing interest in tools that help providers reflect on and improve their communication behaviors, such as the ones developed by ~\cite{wang2024commsense} and~\cite{bedmutha2024conversense}.

Prior work has examined these interpersonal behaviors through the \textit{social signals}. ~\cite{vinciarelli2009social} defines them as \textit{aggregates of verbal and non-verbal behaviors that individuals exhibit towards others within their current social context}. In clinical settings, the most prominent set of behaviors is the Roter Interaction Analysis System (RIAS) (\cite{roter2002roter}), which tracks patients and providers on a set of 21 signals. To obtain fine-grained understanding of the clinical conversations, they recommended breaking visits into \textit{thin slices} of shorter durations (typically 3-5 minutes) and annotate each ``slice'' for all social signals. This requires \textit{expert annotators trained in RIAS} and is time-intensive, therefore \textbf{cannot provide communication feedback at scale}.

Empirically most social signals can be characterized through non-verbal cues such as prosodics and turn-taking (\cite{hartzler2014real}). ~\citet{bedmutha2024conversense} aimed to leverage this association to train models that can track the RIAS signals using such vocalic cues. Despite annotating numerous recordings, they found that the prevalence of high/low levels of social signals was very skewed and in fact they could only model 8 social signals out of the 21. The imbalance is systemic and occurs due to Hawthorne effect described by~\cite{landsberger1958hawthorne}, and may not be ethically addressable with additional data. 
~\citet{chen2024toward} explored the use of text or content of the conversations to fine-tune generalized small language models (SLMs). Despite the limitations, they trained models for each individual task and showed feasibility of tracking 17 out of the 21 signals evaluated across stratified groups of providers. Since the cost of annotating new data is extremely high and the data is highly imbalanced, we envision that tracking these social behaviors requires any model to have \textbf{inherent knowledge} of social and affective behaviors.

While non-verbal cues may offer actionable feedback, their meaning significantly differs across cultures, and audio and video may often be unavailable in real-world deployments. In contrast, historically transcripts have been generated and stored  using manual systems. Recently, automated transcription is rapidly becoming standard through ambient clinical intelligence systems, and as a result, large volumes of clinical conversation transcripts are generated and available. As shown by~\cite{swain2024patients}, transcripts also offer stronger privacy protections compared to raw audio data. This has created an opportunity to infer interpersonal dynamics directly from text alone.

\cite{xu2024mental} and~\cite{yang2023mentalLLaMA} showed how large language models have demonstrated strong capabilities in extracting clinically relevant information from free text, including mental health states such as depression and anxiety. Emerging evidence by~\cite{chen2025detecting} and~\cite{ montiel2024empatheticexchanges} suggests that language models may also capture interpersonal and affective behaviors in other related settings such as therapy. However, it remains unclear whether such models can robustly infer a broad range of social signals, if the performance varies based on known contextual indicators, and whether these predictions can be made reliable enough for deployment.

In this work, we study whether large language models (LLMs) exhibit \textit{inherent capability} to infer social signals from clinical transcripts without supervised training. We evaluate three language models spanning a wide range of scales and architectures, from a small on-device model to a large cloud-hosted model. Across 20 social signals annotated in three-minute transcript segments, we assess zero-shot and few-shot prompting strategies and analyze performance variability across visit stages and patient demographics. Finally, inspired from real-world deployments where query-only models are encouraged and hence uncertainty and confidence scores are unavailable; we investigate the impact of ensembling different model-prompt responses for each task by introducing context-group level agreement probabilities as soft weights. 

Our results show that language models can act as scalable text-based sensors of interpersonal dynamics in clinical conversations, offering a practical pathway for monitoring and studying social signals in real-world healthcare settings.

%% file: paper/2-related-work.tex
\section{Related Work}
\label{sec:related}
Our work builds on prior research on patient-provider communication, social signal processing and the use of large language models in similar tasks.

\subsection{Social Signals in Clinical Conversations}
Patient-provider communication has been widely studied, mostly empirically by \citet{cooper2012associations}, \citet{roter2002roter}, and \citet{hartzler2014real}. \citet{roter2002roter} define 21 social signals through the Roter Interaction Analysis System (RIAS) that impact the success of a visit: *dominance, *attentiveness, *warmth, *engagement, *empathy, *respect, *interactivity, *irritation, *nervousness, **hurriedness, ***sadness, and **emotional distress (*patient \emph{and} provider, **provider only, ***patient only). Recent work by \citet{chen2024toward}, \citet{bedmutha2023ai3}, and \citet{bedmutha2024artificial} has shown that RIAS-inspired social signals may present differently based on patients' race. Additionally, since patient visits often follow a structured flow as described by \citet{robinson2003interactional}, social signal distributions may vary at the start, middle, and end of the visit. These factors motivate examining whether AI performance differs by race and time-within-visit.

Early computational approaches modeled social signals via acoustic, prosodic, and turn-taking features. \citet{hartzler2014real} and \citet{liu2016eqclinic} used audio/video, while \citet{bedmutha2024conversense} and \citet{wang2024commsense} focused on vocalic cues---but these typically addressed few signals and required high-quality recordings. Text-based approaches have gained attention with increasing availability of clinical transcripts; \citet{chen2024toward} showed that fine-tuned small language models could predict up to seventeen RIAS signals. However, these methods rely on supervised, task-specific training, limiting scalability. To date, no prior work has examined whether LLMs, as foundation models without fine-tuning, can infer a broad range of social signals from text alone.

Building reliable systems remains challenging due to labor-intensive annotation, signal sparsity, and limited generalizability of institution-specific models. Yet the growing ubiquity of AI scribes, as noted by \citet{tierney2024ambient}, and privacy-aware sensing systems, such as the ones developed by \citet{bedmutha2023privacy} and \citet{boovaraghavan2024kirigami} suggest transcripts will soon be widely available. \citet{bedmutha2024artificial} and \citet{bascom2024designing}, have emphasized that providers value timely, personalized feedback, pointing to the need for unified systems that operate without extensive retraining.

%
%

\subsection{Large Language Models for Patient-Provider Communication}
In healthcare, LLMs have shown promise for clinical documentation and mental health assessment, as demonstrated by \citet{xu2024mental} and \citet{yang2023mentalLLaMA}. For patient-provider communication specifically, \citet{ayers2023comparing} found that LLM-generated responses can be perceived as accurate and empathic in asynchronous settings, while \citet{vishwanath2024role} and \citet{tu2025towards} have explored LLMs for patient education and conversational agents.

Recent studies suggest LLMs can reason about affective properties of health-related text. \citet{luo2024assessing} showed LLMs can assess empathy in patient-provider messages, and \citet{feng2023affect} demonstrated emotional state capture in therapeutic conversations; outside healthcare, \citet{lei2023instructerc} found strong performance on emotion recognition in dialogue. 

However, it remains unclear whether these capabilities extend to naturalistic, synchronous clinical conversations involving role asymmetries and subtle social cues; or whether performance varies across patient populations, visit stages, and signal types.



%% file: paper/3-methods.tex
\section{Modeling Social Behavior in Clinical Conversations}
\label{sec:methods}

We frame social signal processing (SSP) as the task of inferring interpersonal behaviors expressed in short segments of clinical conversations or slices. Because many social behaviors lack precise definitions and exhibit substantial annotator subjectivity, SSP can also be interpreted as a human-AI alignment problem, where a correct prediction reflects agreement with expert human judgments rather than an objective ground truth.

\subsection{Dataset and Task}
\label{sec:dataset}
We evaluate SSP using the Establishing Focus (EF) dataset from~\citet{AHRQ2006}, which contains real-world primary care visits from the West Coast of the United States, annotated with RIAS-inspired global affect ratings as defined by~\citet{roter2002roter}. This is the largest annotated dataset and used widely for empirical (such as~\cite{vannoy2011suicide, bascom2024designing}) and quantitative studies(such as~\cite{bedmutha2024conversense, zolensky2025speaker})

Following the thin-slice methodology of \citet{roter2011slicing}, each visit is segmented into independent three-minute ``thin slices'', yielding 512 annotated segments across 91 visits conducted by 22 providers. The audio recordings are transcribed using Whisper-v3-large~\citep{radford2022whisper} and diarized using pyannote~\citep{bredin2020pyannote}, which have become the standard for conversational affective tasks~\citet{agrawal2025seamless}. 
After preprocessing for empty slices, our final dataset consists of 508 high-quality slices. The coding scheme includes 21 social signals attributed to the patient, provider, or both. These are grouped into:

\begin{enumerate}[leftmargin=*]
\item \textit{Type-I Signals:} Dominance, attentiveness, warmth, engagement, empathy, respect, interactivity, and hurriedness. All are rated for both patient and provider except hurriedness (provider only). Scores range from 1 (low) to 6 (high), with 3.5 as the neutral midpoint.
\item \textit{Type-II Signals:} Irritation, nervousness, sadness, and emotional distress; sadness and emotional distress are patient-only. These capture negative affective states, where 1 denotes absence and ratings increase with expression up to 6.
\end{enumerate}

\begin{figure}[t]
\centering
\includegraphics[width=.9\linewidth]{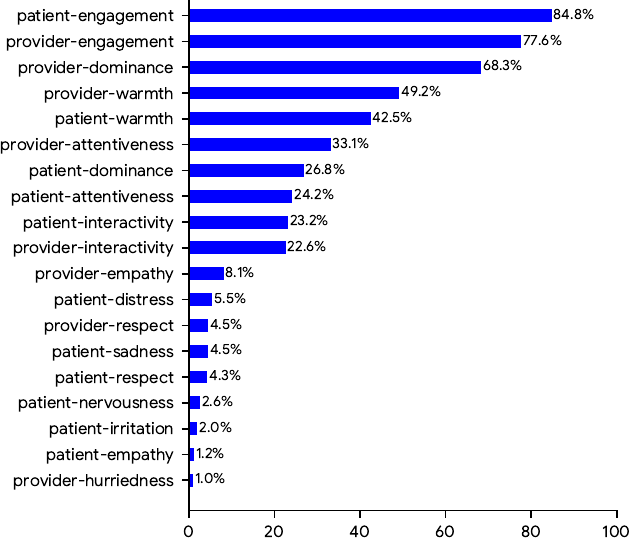}
\vspace{-1.2em}
\caption{Percentage of \textit{high} social signal labels distribution across 3-minute segments across the entire sample}
\label{fig:signal_distribution}
\vspace{-1.5em}
\end{figure}

One signal (provider nervousness) is never observed, leaving 20 signals for analysis. Following prior work by~\citet{chen2024toward} and ~\citet{bedmutha2024conversense}, we binarize all labels. For Type-I signals, we predict whether behavior is expressed at a higher-than-average level (threshold 3.5). For Type-II signals, we predict presence versus absence (threshold 1.5). This formulation reflects both limited label granularity and the extreme sparsity of negative or disruptive behaviors in clinical encounters. The final data distribution (\autoref{fig:signal_distribution}) yields a binary prediction task across 20 behaviors per 3-minute slice of conversation.

\paragraph{Present-day utility:}~\cite{bedmutha2024artificial} collected another dataset of 108 visits in 2023 consisting of a diverse representation of LGBTQ+ and BIPOC participants. They trained social signal models on the EF dataset using primitive audio features and showed that these models directly generalized on their new dataset in 5/7 signals and all models beat the fairness baselines. This instills confidence that findings from our current analysis have the potential to translate to present-day communication patterns.

\subsection{Evaluating Feasibility of SSP}
We describe below the overview of our feasibility benchmarking to explore if at all language models can track affective information from clinical conversations. The EF dataset exhibits severe class imbalance across nearly all signals (Figure~\ref{fig:signal_distribution}), particularly for Type-II behaviors, which are rare or nearly absent. Moreover, positive instances are often clustered within a small number of visits or providers, limiting diversity and making supervised learning unstable. These challenges stem from known biases in observational healthcare data, including sampling bias and the Hawthorne effect. As a result, training task-specific models from scratch or via fine-tuning is infeasible for many signals. Instead, we investigate whether large language models can leverage internalized social and conversational knowledge to perform SSP in zero- and few-shot settings, without relying on extensive labeled data. 

\subsubsection{Prompting and Models}
Each social signal is treated as an independent binary classification task applied to each transcript slice. We evaluate standard prompting strategies including zero-shot, few-shot, and chain-of-thought prompting. Prompts specify the model's role, the target social signal, and the required output format. Further details of the prompts are provided in Appendix~\ref{app:prompts}.

We evaluate three representative language models spanning a wide range of scales: a small instruction-tuned encoder-decoder model (Flan-T5-Base), a mid-sized decoder-only model (Gemma2-2B), and a large-scale decoder-only model (Llama 3.1 405B). All models are run locally to ensure data privacy. Flan and Gemma are used in \textit{generation} while Llama is used in \textit{instruct} mode.

\paragraph{Clinical Feasibility:} 
The first two models represent on-device deployment; the latter simulates cloud inference. Our health system's protocols restricted access to logit distributions, so we operate under the constraint that large models are \textbf{query-access only}. Despite the binarized state, tracking changes in these behaviors can support various reflective visualizations for providers~\cite{bedmutha2024conversense}.

\subsubsection{Evaluation}
Each \texttt{<model, prompt>} pair is treated as a distinct configuration. Models are run deterministically using greedy decoding (temperature zero), ensuring consistent responses and removing the need for resampled evaluations. When applicable, class predictions are derived via logit comparison between binary labels. Some prompt-model combinations were excluded due to systematic output failures and are discussed in Section~\ref{sec:limitations}.

As a supervised baseline, we reproduce the BERT-based fine-tuned models from~\citet{chen2024toward}, using leave-one-provider-group-out cross-validation and class-weighted loss. We varied the learning rate (1e-5 to 6e-5), loss parameter($\gamma=\{0,1,2\}$) and weight decay ($\{0, 0.01, 0.1\}$). Across all the configurations, due to instability caused by extreme imbalance, signals with high variance (\textgreater0.2) across folds are treated as non-modelable. 

Finally, we train a lightweight logistic regression ensemble over all LLM configurations, to test whether diverse  model-prompt configurations yield complementary predictions.

\begin{table*}[t]
\centering

\resizebox{\textwidth}{!}{
\begin{tabular}{lccccccccccc}
\toprule
\textbf{Social Signal} & \textbf{BERT} & \textbf{FLAN-T5} & \textbf{FLAN-T5} & \textbf{Gemma} & \textbf{Gemma} & \textbf{Gemma} & \textbf{LLaMA} & \textbf{LLaMA} & \textbf{LLaMA} & \textbf{LLaMA} & \textbf{Ensemble} \\
 & \textbf{(fine-tuned)} & \textbf{ZS} & \textbf{FS} & \textbf{ZS} & \textbf{FS} & \textbf{COT} & \textbf{ZS} & \textbf{FS} & \textbf{COT} & \textbf{FSCOT} & \textbf{(LOGO)}\\
\midrule
\textit{Type-I Signals} & \\
Provider Dominance & \textbf{0.618} & \underline{0.549} & 0.539 & 0.522 & 0.502 & 0.478 & 0.507 & 0.498 & 0.538 & 0.495 & 0.603 \textsubscript{(0.13)}\\
Provider Attentiveness & \textbf{0.563} & 0.531 & 0.520 & 0.509 & 0.489 & 0.523 & 0.463 & \underline{0.536} & 0.437 & 0.523 & 0.624 \textsubscript{(0.09)}\\
Provider Warmth & \textbf{0.571} & 0.467 & 0.469 & 0.518 & 0.481 & 0.506 & 0.517 & \underline{0.545} & 0.512 & 0.511 & 0.558 \textsubscript{(0.10)}\\
Provider Engagement & 0.543 & 0.489 & 0.489 & 0.459 & 0.483 & 0.500 & 0.513 & \underline{0.556} & 0.534 & \textbf{0.564} & 0.609 \textsubscript{(0.19)}\\
Provider Empathy & 0.500 & 0.530 & 0.532 & 0.516 & 0.579 & 0.489 & 0.610 & \underline{0.623} & 0.585 & \textbf{0.629} & 0.628 \textsubscript{(0.13)}\\
Provider Respect & N/A & 0.494 & 0.499 & 0.509 & 0.532 & \textbf{0.552} & \underline{0.542} & 0.512 & 0.529 & 0.495 & 0.488 \textsubscript{(0.06)}\\
Provider Interactivity & \textbf{0.643}  & 0.510 & 0.519 & 0.499 & 0.512 & 0.512 & 0.530 & 0.539 & \underline{0.560} & 0.542 & 0.484 \textsubscript{(0.06)}\\
Patient Dominance & 0.583 & 0.524 & 0.516 & 0.471 & 0.491 & 0.500 & \underline{0.601} & 0.554 & \textbf{0.608} & 0.568 & 0.587 \textsubscript{(0.08)}\\
Patient Attentiveness & 0.549 & 0.520 & 0.518 & 0.496 & 0.476 & 0.536 & 0.516 & \textbf{0.577} & 0.515 & \underline{0.576} & 0.589 \textsubscript{(0.03)}\\
Patient Warmth & \textbf{0.539} & 0.473 & 0.477 & 0.488 & 0.510 & 0.505 & \underline{0.577} & 0.557 & 0.547 & \textbf{0.582} & 0.564 \textsubscript{(0.06)}\\
Patient Engagement & \textbf{0.614} & 0.549 & 0.548 & 0.462 & 0.513 & 0.480 & 0.553 & \underline{0.599} & 0.557 & 0.595 & 0.554 \textsubscript{(0.04)}\\
Patient Empathy & N/A & 0.566 & 0.570 & 0.393 & 0.483 & 0.491 & 0.456 & 0.487 & \textbf{0.589} & \underline{0.575} & 0.464 \textsubscript{(0.18)}\\
Patient Respect & \textbf{0.599} & 0.490 & 0.499 & 0.514 & 0.413 & 0.529 & 0.496 & \underline{0.546} & 0.513 & 0.495 & 0.575 \textsubscript{(0.10)}\\
Patient Interactivity & \textbf{0.705} & 0.540 & \underline{0.544} & 0.454 & 0.442 & 0.515 & 0.492 & 0.488 & 0.500 & 0.480 & 0.599 \textsubscript{(0.13)}\\
Provider Hurriedness & N/A & 0.350 & 0.358 & \underline{0.555} & 0.446 & \textbf{0.568} & 0.551 & 0.492 & 0.267 & 0.448 & 0.664 \textsubscript{(0.24)}\\
\midrule
\textit{Type-II Signals} & \\
Provider Irritation & N/A & 0.585 & 0.457 & 0.550 & 0.125 & 0.562 & 0.857 & \textbf{0.947} & \underline{0.927} & 0.734 & 0.901 \textsubscript{(0.18)}\\
Patient Irritation & N/A & 0.520 & 0.527 & 0.547 & 0.465 & 0.362 & \textbf{0.685} & \underline{0.631} & 0.583 & 0.612 & 0.659 \textsubscript{(0.19)}\\
Patient Nervousness & 0.600 & 0.492 & 0.498 & 0.404 & 0.489 & 0.379 & 0.514 & \textbf{0.726} & 0.586 & \underline{0.608} & 0.650 \textsubscript{(0.23)}\\
Patient Sadness & 0.600 & 0.545 & 0.550 & 0.516 & 0.469 & 0.461 & 0.624 & 0.686 & \underline{0.696} & \textbf{0.711} & 0.681 \textsubscript{(0.15)}\\
Patient Distress & 0.500 & 0.525 & 0.515 & 0.508 & 0.549 & 0.441 & 0.568 & \textbf{0.659} & \underline{0.639} & 0.633 & 0.644 \textsubscript{(0.10)}\\
\midrule
MEAN & \underline{0.581} & 0.512 & 0.507 & 0.494 & 0.473 & 0.494 & 0.559 & \textbf{0.588} & 0.561 & 0.569 & 0.606 \\
STD & 0.05 & 0.05 & 0.04 & 0.04 & 0.09 & 0.05 & 0.09 & 0.11 & 0.12 & 0.07 & 0.09 \\
\bottomrule
\end{tabular}
}

\vspace{1em}

\caption{Table shows the overall performance of different models and configurations as balanced accuracies. We use fine-tuned BERT~\citep{chen2024toward} for comparison with LLM predictions. We also train an ensemble model based on the predictions of LLMs over groups of providers and report additive effects in performance. Values in \textbf{bold} are the best-performing model and \underline{underline} refers to the second-best model. In binary settings, a balanced accuracy greater than 0.5 implies the model performs better than chance}
\vspace{-2em}
\label{tab:overall_acc}

\end{table*}

%% file: paper/4-analyses.tex
\section{Analyses}
\label{sec:analysis}
We evaluate language models as text-based sensors of social behavior by comparing their predictions against expert annotations across 20 social signals. Due to extreme class imbalance across nearly all signals, we report \emph{balanced accuracy}, which equally weighs performance on majority and minority classes and is standard for behavioral sensing tasks with skewed label distributions such as the ones showed by~\cite{xu2024mental} and~\cite{kaufman2025predicting}.

\subsection{Overall Performance Under Severe Imbalance}
Across all model–prompt configurations, social signal prediction is fundamentally constrained by data imbalance making fine-tuning impractical for many signals. In contrast, prompted large language models achieve above-chance balanced accuracy for \textbf{all 20 social signals} without any task-specific training. While absolute performance varies across signals, these results indicate that LLMs encode prior knowledge about interpersonal dynamics that can be elicited through prompting. Many signals, particularly Type-II affective states such as irritation, nervousness, and emotional distress, are rare or nearly absent in the dataset. However LLMs perform strongly despite these imbalances. While the large LLaMA with few-shot prompting provides best performance over all the subtasks, there is \textbf{no clear one-configuration-fits-all} for all the subtasks. This is further highlighted when the naive-ensemble betters most individual models.

\subsection{LLM behavior across patient race}
\label{sec:analysis_fair}
Since social signals are closely linked to patient race, often due to providers' implicit biases as shown by~\cite{cooper2012associations} and~\cite{bedmutha2023ai3}, we examine whether the LLM performance varies between White and non-White patients. This formulation allows us to frame this evaluation as a fairness investigation.

\paragraph{(1) Performance Gap between White and non-White patients:}
Our dataset contains 74 visits involving White patients and 17 involving non-White patients. Despite differences in label distributions between the groups, we observe that most balanced accuracy scores (computed over all the configurations) across both groups remain above 0.5 but show substantial predictive disparity. Only five social signals scored below 0.5, indicating that in general LLMs tend to be able to track social signals for both demographic subgroups. However, certain signals were entirely absent in the non-White group, which likely contributes to the larger observed performance gaps for example, \textit{provider hurriedness} ($\overline{\triangle} = 0.235$) and \textit{provider irritation} ($\overline{\triangle} = 0.130$). Interestingly, even when the label distributions were similar, we found that LLMs were better able to detect patient empathy among non-White patients ($\overline{\triangle} = -0.110$), suggesting possible unintended biases in model attention or interpretation. Table~\ref{tab:fair} in the Appendix shows complete results.

\begin{figure*}
\vspace{-1em}
\centering
\includegraphics[width=.77\linewidth]{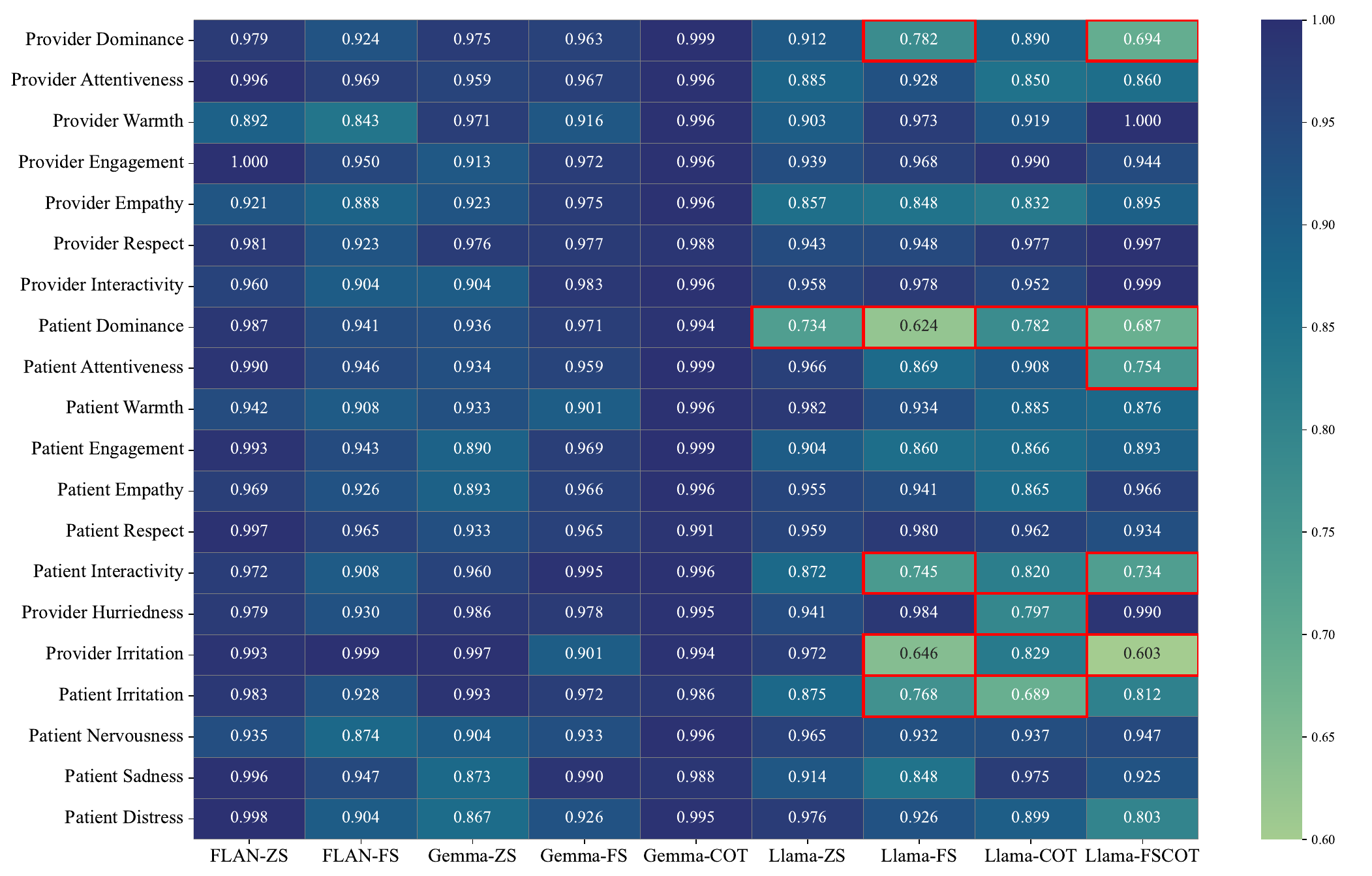}
\vspace{-1em}
\caption{Demographic Parity Ratio between white (n=74) and non-white (n=17) patients. Most configurations follow the four-fifths rule (threshold = 0.8), but some disparity was observed for LLaMA.}
\vspace{-2em}
\label{fig:dpr}
\end{figure*}

\paragraph{(2) Perceived Fairness of Individual Model Predictions:}
We investigate the fairness of the model predictions through Demographic Parity Ratio between White and non-White patients. We apply the four-fifths rule as a threshold, which indicates disparate impact occuring if the performance ratio between the groups falls below 4/5. The parity ratios are presented in Figure~\ref{fig:dpr}. We find very few instances of disparity: FLAN-T5 and Gemma configurations show no evidence of disparate impact, while Llama exhibits disparities on six social signals—specifically, \textit{provider dominance, patient dominance, patient interactivity, provider hurriedness, provider irritation, and patient irritation}. These results suggest that Llama may be sensitive to linguistic cues that correlate with patient demographics which are often linked to broader social determinants of health.

\subsection{Does performance shift across stages of a medical visit?}
\label{sec:analysis_segments}
Clinic visits usually follow a predictable flow.~\citet{robinson2003interactional} described this as a series of phases: opening (greetings, patient concerns), middle (detailed symptom discussion and diagnosis), and closing (summarization and treatment planning). Since each segment has a set pattern in conversations, the LLM performance may vary and be indicative of the underlying content. We divide each visit into \textbf{three segments: the start (first three-minute slice), middle (all intermediate slices), and end (last slice)}. Our subsequent analysis explores this variation in manually-coded behaviors and model predictions:

\begin{figure*}[h]
\vspace{-1em}
\centering
\includegraphics[width=.9\linewidth]{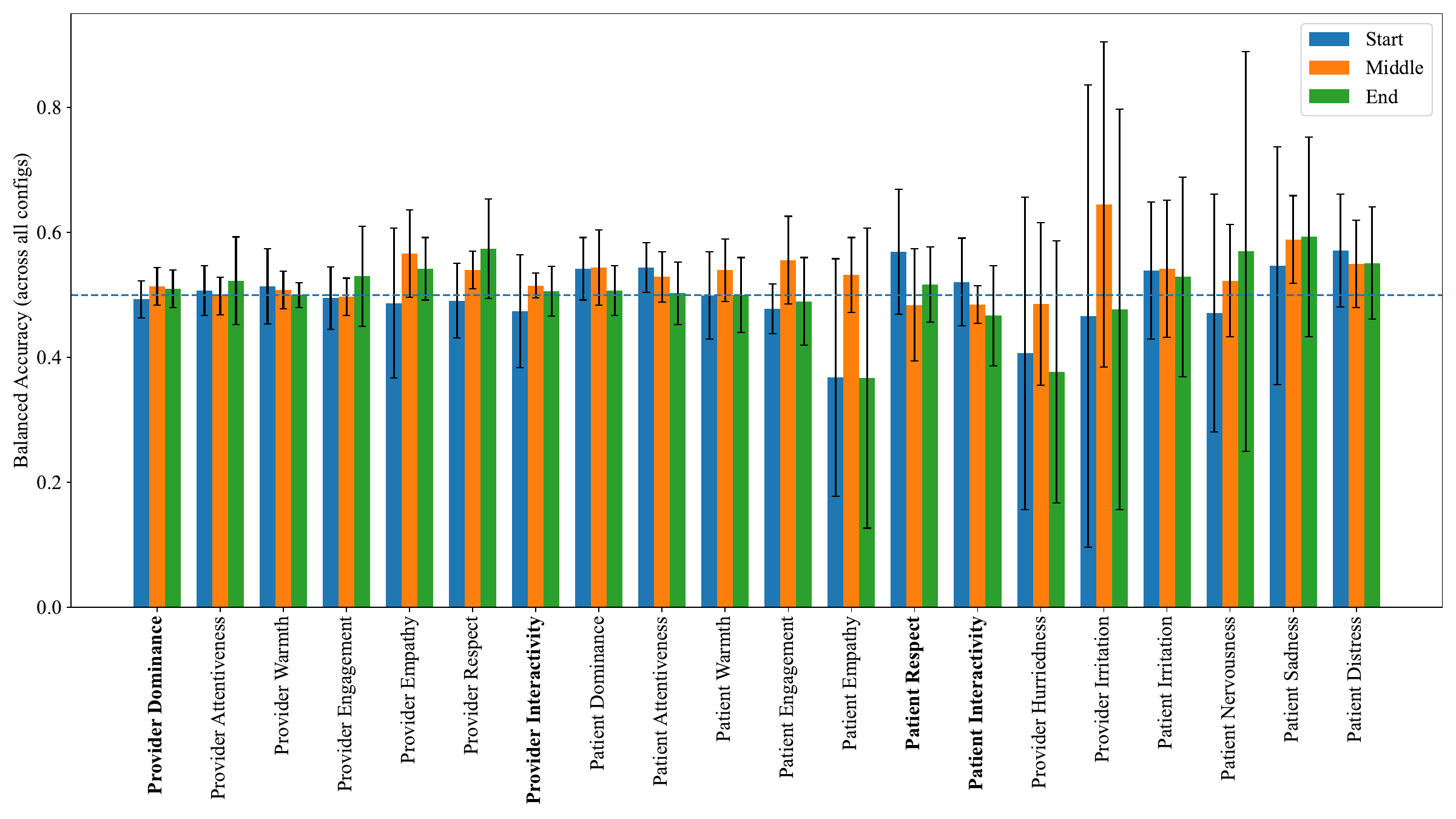}
\vspace{-1em}
\caption{Figure shows the variability in performance at different segments (start, middle, end) of a visit. The bars represent average performance over all the model configurations and standard deviations describe the variability within the models. Signals marked in \textbf{bold} describe the ones whose distribution in a visit varies significantly over the segments}
\label{fig:acc_time}
\vspace{-1em}
\end{figure*}

We examine whether the distribution of binary social signal labels differs across three \textit{segments} of each visit: start, middle, and end. For each segment, we compute balanced accuracy across all configurations (Figure~\ref{fig:acc_time}). Language models perform \textbf{least accurately at the start}, achieve \textbf{highest performance in the middle}, and show a modest decline toward the end; likely because the middle involves more balanced interactive exchanges that provide richer contextual cues for LLM-based social behavior inferences.

%% file: paper/5-ensemble.tex
\section{Post-Hoc Ensembling by Contextual Aggregation}
\label{sec:ensemble}

Section~\ref{sec:analysis} showed that LLMs demonstrate strong inherent capability for inferring social signals from clinical transcripts, often matching or exceeding fine-tuned task-specific models. However, performance varies substantially across signals, patient race, and temporal segments of a visit, and no single configuration consistently dominates.  Therefore real-world deployment will be limited by task-specific and context-dependent gaps in performance. This motivates ensemble-based inference across multiple model-prompt configurations to attain robustness.

In clinical deployments, models are often accessed via APIs exposing only discrete predictions without probabilities. Most approaches of uncertainty or confidence estimation rely on underlying disribution~\citep{huang2024survey} therefore soft-voting is infeasible. We therefore propose an agreement-weighted ensemble that operates under query-only access and leverages auxiliary metadata available at inference time.

\subsection{Assumptions}
We make these assumptions about our setting.
\begin{enumerate}[leftmargin=*]
\item \textbf{Query-access to configurations:} Multiple models and prompts can be queried, but only discrete predictions are returned.
\item \textbf{Auxiliary information:} Patient race and transcript position within the visit (start, middle, end) are available at inference time.
\item \textbf{Benchmark statistics:} A human-annotated benchmark provides performance estimates for each configuration (Table~\ref{tab:overall_acc}).
\item \textbf{Dataset imbalance:} Label rarity and demographic imbalance may influence reliability of performance estimates.
\end{enumerate}

\subsection{Group-Level Agreement Estimation}
From Section~\ref{sec:analysis}, it is evident that benchmark performance for a task or subgroup (such as race or time-of-visit) can be misleading due to disparities in the dataset. Therefore, we will have less confidence in a configuration that performs suboptimally or a group that is less represented in the benchmark. Thus for each group in the known dataset, if we can weigh individual samples by their perceived extent of \textbf{agreement with ground truth}, the ensemble will aim to achieve the best performance while being aware of the current capabilities of the model-prompt configurations available. Note, in subsequent description, we refer to each configuration as an individual \textit{model}.

Conventional estimates of agreement, uncertainty, or confidence typically rely on probability outputs, as reviewed by~\cite{huang2024survey}. However, \citet{chen2023quantifying} showed that uncertainty can alternatively be estimated by evaluating consistency across prompts or model variations. Inspired by this perspective, we estimate net agreement within groups conditioned on available auxiliary information.

Let $x_0$ denote transcript input, $y$ the true label, $\hat{y}$ the model prediction, and $x_1,\dots,x_n$ auxiliary variables defining groups. Agreement is defined as:

\vspace{-1.5em}
\begin{multline}
\Pr(\text{agree} \mid x_{0}, \ldots, x_{n}; \theta) = \\ 
\sum_{k=0}^{K} \Pr(\hat{y}=k \cap y=k \mid x_{0}, \ldots, x_{n}; \theta)
\end{multline}

\noindent
Using the chain rule:
\begin{equation}
\begin{aligned}
&\Pr(\text{agree} \mid \mathbf{x}_{0:n}; \theta) = \sum_{k=0}^{K} 
\underbrace{\Pr(\hat{y}=k \mid y=k, \mathbf{x}_{1:n}; \theta)}_{\text{model precision for group}} \\
&\quad \times \underbrace{\Pr(y=k \mid \mathbf{x}_{1:n})}_{\text{label prevalence}}
\times \underbrace{\Pr(\mathbf{x}_{1:n} \mid x_0)}_{\text{group prevalence}}
\end{aligned}
\label{eq:agreement}
\end{equation}

The first term reflects model precision conditioned on group membership, while the remaining terms capture label imbalance and subgroup prevalence. Agreement therefore accounts for both task imbalance and demographic disparities.

\subsection{Agreement-Weighted Ensemble Construction}

We consider binary classification ($K \in \{-1,1\}$). Auxiliary variables include race ($X_1 \in \{\text{White}, \text{Non-White}\}$) and visit segment ($X_2 \in \{\text{Start}, \text{Middle}, \text{End}\}$), yielding six mutually exclusive groups.

For each training sample $l$, model $\theta_m$ outputs $\theta_m(x_0^l)\in\{-1,1\}$. Agreement weights $w_{g,m}$ are estimated per model and group using Eq.~\ref{eq:agreement}. Each sample inherits its group weight ($w_{l,m}=w_{g,m}$), and scaled predictions are computed as
$s_{l,m}=w_{l,m}\cdot\theta_m^l$. An aggregator function $f$ is trained on scaled scores across models. Algorithm~\ref{alg:training} describes the approach stepwise.

\begin{algorithm}[h]
\caption{Ensemble Training for each task}
\label{alg:training}
\KwIn{Training data $\{(x_0^l, x_1^l, x_2^l, y^l)\}_{l=1}^N$, models $\{\theta_m\}_{m=1}^M$}
\KwOut{Group-level weights $w_{g,m}$}

Define groups $\mathcal{G} = \{\text{White}, \text{Non-White}\} \times \{\text{Start}, \text{Middle}, \text{End}\}$\;

\ForEach{$g \in \mathcal{G}$}{
    \For{$m \leftarrow 1$ \KwTo $M$}{
        Collect predictions $\theta_m(x_0^l)$ for samples with $(x_1^l,x_2^l)\in g$\;
        Compute agreement $w_{g,m}$ using Eq.~\ref{eq:agreement}\;
    }
}

Fit aggregator $f$ on $\{s_{l,m}=w_{g(l),m}\theta_m(x_0^l)\}$\;
\Return{$\{w_{g,m}\}$ and $f$}\;
\end{algorithm}

The inference is similarly driven, by first assigning a slice to a group and then invoking the $f$ for that setting.





\paragraph{Experimental Setup}
We evaluate using leave-one-group-out cross-validation across five provider groups to avoid leakage. Aggregators include Logistic Regression and Decision Trees (max depth=2). We vary top-$k$ models ($k=2,3,5,9$ [all]) ranked by benchmark performance. A naive ensemble baseline uses identical weights ($w=1$).

\subsection{Ensembled SSP Performance Evaluation}
We describe the effect of aggregation using three approaches for all the 20 signals: (i) Raw or individual best performing model, (ii) LOGO, a leave-one-group-out ensemble baseline using logistic regression without any group-level adjustments (i.e. $w = 1$ for all inferences), and (iii) Ensemble (Best) representing the best-performing configuration of agreement-weighted ensembles. The raw scores describe the idea of choosing the only best known model in the benchmark for the task. LOGO ensembles over all the models but unaware of the group-level (auxiliary) information and the Ensemble scores show the effect of group-level aggregation. 

\begin{figure*}[t]
\vspace{-1em}
\centering
\includegraphics[width=.9\linewidth]{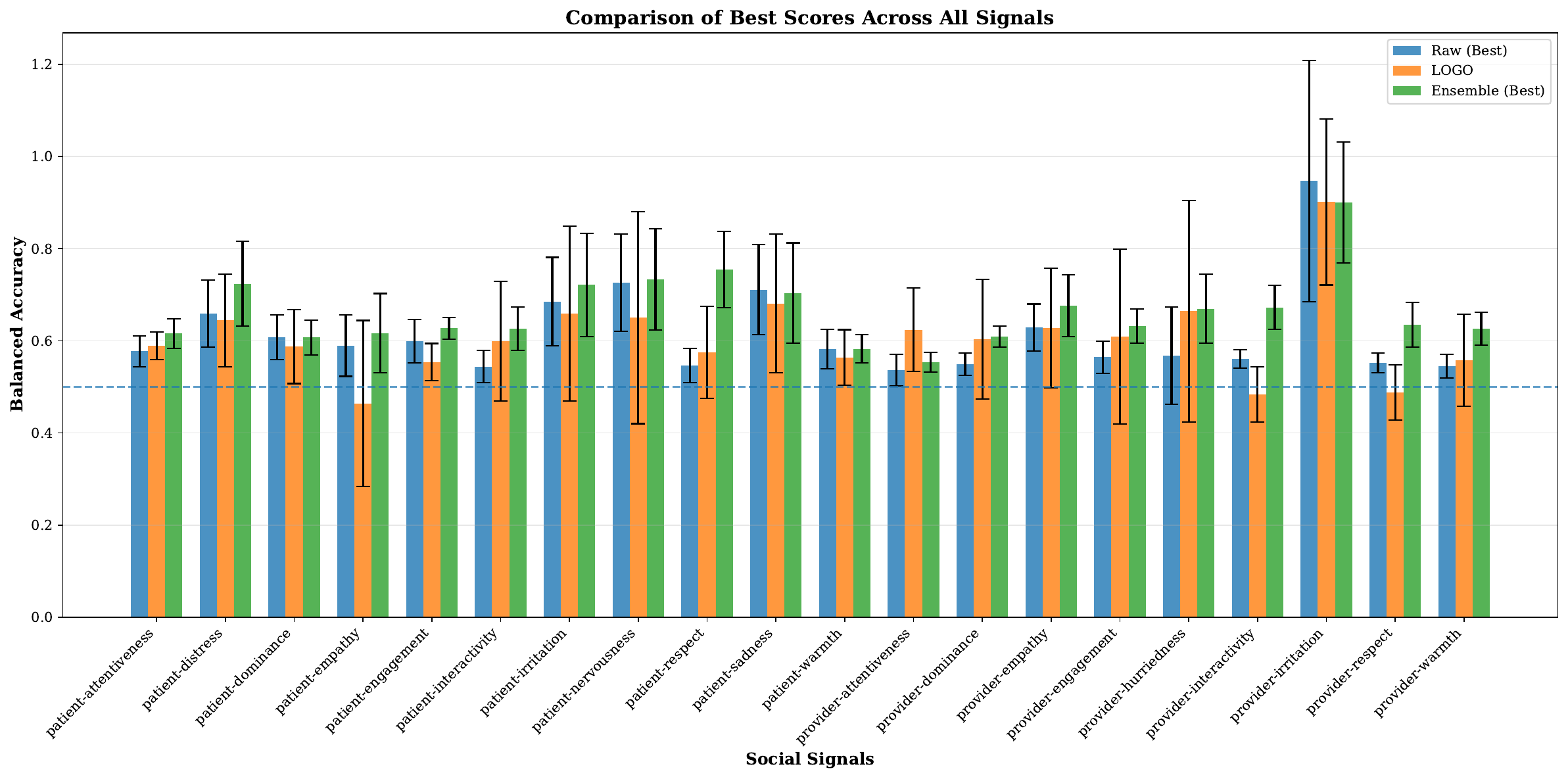}
\vspace{-1em}
\caption{Comparison of balanced accuracy across all 20 social signals for three methods: Raw (Best) shows the best-performing individual model per task, LOGO represents a leave-one-group-out logistic regression ensemble without group-level weighting, and Ensemble (Best) is our agreement-weighted ensemble approach. Error bars indicate standard deviation.}
\label{fig:ensemble_comparison}
\end{figure*}

Figure~\ref{fig:ensemble_comparison} shows balanced accuracy across all social signals for the three approaches. Averaged over 20 signals and folds, the mean balanced accuracy for Raw (Best) is 0.614 (SD = 0.063), for LOGO is 0.606 (SD = 0.124), and for Ensemble (Best) is 0.664 (SD = 0.063). 

\paragraph{Aggregation is inherently beneficial.} On comparing the \textsc{Raw} and \textsc{LOGO} we observe that even even simple aggregation enhances performance for most signals. The \textsc{LOGO} ensemble matches or exceeds the best individual model, with particularly notable gains for \textit{patient attentiveness}, \textit{patient dominance}, and \textit{provider hurtfulness}, where ensemble averaging clearly stabilizes and improves predictions.

\paragraph{Group-weighted aggregation provides substantial improvements.} With the use of added contextual information, the \textsc{Ensemble} approach a mean balanced accuracy of 0.664 an 5\% increase over the best single model (Llama-FS) and a 5.8\% improvement compared to \textsc{LOGO}. In Figure~\ref{fig:ensemble_comparison}, in fact we note that in 16 out of 20 signals, the group-weighted aggregation results in the best performing model. This improvement is achieved by maintaining comparable if not lower deviation over folds.

These results demonstrate that while aggregation is inherently valuable, \textit{how} we aggregate matters significantly. By leveraging contextual information (race, time-within-visit), the weighted ensemble selectively amplifies model strengths while maintaining stability—achieving best performance on 16 of 20 signals without increased variance and in fact reduces variance in 10 among them. This makes agreement-weighted ensembling particularly suitable for clinical deployment where reliability across diverse patient populations is essential.

%% file: paper/6-discussion.tex
\section{Discussion}
\label{sec:discussion}
In this work, we introduced social signal processing (SSP) for clinical conversations using LLMs, demonstrating feasibility across 20 signals and proposing contextual ensembling to address performance variability. Since the best performing configuration is very different for each task, we also note that SSP deployment is a benchmark-sensitive space. We discuss implications of our findings below.

\subsection{LLM Orchestration for Clinical SSP}
We know that every model tends to have varying degrees of performance and precision for each task. From our experiments in aggregation (Section~\ref{sec:ensemble}), we observed that even within the three pipelines described, the best performing individual models and parameters of aggregation may vary based on factors such as the task, group (race, time-within-visit), number of models to be included (top-k), and aggregator (logistic regression or decision tree). 

Additionally, on-device model may fail, or cloud models may experience downtime; an intelligent system could delay feedback for reliability, or provide immediate but less reliable feedback depending on context. Since we now have extensive information of the best performing model for each user, task and model, feedback systems can leverage task-specific insights on the ideal choice of parameters, base-models and prompts to personalize how to best address the users' needs. Therefore, our comparative benchmarking (Table~\ref{tab:overall_acc}) and aggregation experiments (Appendix~\ref{appendix:ensembling}, Table~\ref{tab:highest_scores_combined}), can help interface designers route feedback through the ideal configurations, in such a way to meaningfully aggregate their results; ultimately this will allow them to design an optimal orchestration layer of LLM systems for SSP tasks.

\subsection{Toward Trustworthy LLM Feedback}
Understanding when the model is likely right (and when it may be wrong) is essential. When LLM outputs are shown as communication feedback~\citep{steenstra2025scaffolding}, or used to score communicative behavior~\citep{la2025exploring}, designers need a sense of reliability. As~\cite{fargen2012observational} noted, unchecked alerts can cause notification fatigue, especially for clinicians. \cite{park2023advancing} showed that selective confidence-based feedback can also improve trust and adoption.

While group-agreement scores (\autoref{eq:agreement}) provide a coarse confidence measure, linguistic cues could enable finer-grained ``difficulty scores'' for more nuanced confidence estimation.

Such a metric could help inform whether feedback should be pushed to a user, delayed, or adapted ultimately supporting safer and more trustworthy integration of LLM-based feedback systems into high-stakes environments like clinical care.

\subsection{Limitations and Future Work}
\label{sec:limitations}
While our work demonstrates the promise of LLMs in modeling social behaviors in clinical conversations, several important limitations must be acknowledged. These constraints highlight opportunities for further research and refinement, and point to the necessary steps toward deploying social signal processing (SSP) systems that are robust, fair, and clinically useful.

First, we conducted our experiments on an annotated subset of the \emph{Establishing Focus (EF)} dataset, which, although valuable, was collected nearly two decades ago. The dataset lacks a formal, codified annotation guide for its 21 social signals, relying instead on a \textit{trainer-trains-trainer} method where new annotators learned from predecessors until sufficient agreement was reached. This introduces potential label drift and limits our ability to design prompt strategies around standardized definitions---something more feasible in aligned domains such as motivational interviewing~\citep{steenstra2025scaffolding, shah2022modeling} or depression detection~\citep{chen2024depression}. Nevertheless, the EF dataset remains one of the few with real-world, RIAS-style multi-signal annotations, making it a valuable starting point.

A more critical limitation stems from the high class imbalance across most social signals. Several affective states—such as \textit{provider nervousness}—were either completely absent or extremely rare, forcing us to recast the task as a binary classification problem. This imbalance is likely exacerbated by the Hawthorne effect~\citep{landsberger1958hawthorne}.  Promising alternatives include data from standardized patients, as done by~\cite{bedmutha2024conversense}, which we are exploring to elicit difficult conversations in realistic and controlled settings.

From a modeling perspective, we encountered generation failures in smaller models. For instance, FLAN-T5-Base failed to produce coherent responses in Chain-of-Thought prompting, and Gemma2-2B struggled in FS-CoT configurations. Although their raw logits could be extracted for classification, their outputs lacked full interpretability and consistency, making qualitative or explanation-based evaluation impractical. These limitations highlight the importance of model expressiveness—not just accuracy—in sensitive behavioral tasks.

In formative fine-tuning experiments, we found that larger models often overfitted to the majority class, especially when constrained by smaller batch sizes on limited hardware. Despite efforts using dropout, LoRA (\cite{hu2022lora}), and frozen encoders, fine-tuned performance did not surpass zero- or few-shot prompting in a meaningful way. This informed our decision to focus on maximizing performance via prompting, and to treat LLMs as \emph{general-purpose social interpreters} rather than narrow classifiers. As more datasets emerge, our ongoing work will also include fine-tuning medium sized models.

%% file: paper/7-conclusion.tex
\section{Conclusion}
\label{sec:conclusion}
Interpersonal communication is central to quality care, yet assessing social behavior at scale is difficult. We present a no–fine-tuning pipeline that extracts 20 affective social signals directly from clinical transcripts. Across three models and four prompting strategies, many configurations remained reliable despite severe label imbalance. Larger models benefited most from prompt augmentation, while smaller models matched or exceeded performance on specific tasks. Performance shifted across patient race and visit segment, with LLaMA showing disparities on six signals while FLAN–T5 and Gemma showed none.

To address variability under realistic constraints, we propose: under assumption of a query-access system, an agreement-weighted ensemble that uses demographic- and segment-specific agreement weights. This approach improved balanced accuracy by 5\% over the best single model and 5.8\% over an unweighted ensemble, outperforming the best individual model on 16/20 signals while requiring only binary outputs. Overall, this work demonstrates a practical method for reliable, transcript-based social signal modeling in clinical care.

\section{Ethical Implications and Responsible Usage}
Our work uses real-world clinical recordings annotated by expert raters; biases due to patient and score imbalances may impact findings. We maintained a HIPAA-compliant environment, and any deployment should ensure end-to-end data security. Finally, while sensing communication skills benefits providers, such tools could be misused for surveillance by administrators—data privacy for each inference is critical.


%% file: paper/A-Prompt.tex
\counterwithin{table}{section}      
\renewcommand{\thetable}{\thesection\arabic{table}} 

\newpage
\onecolumn 
\section{Prompt Creation}
\label{app:prompts}
We describe below the prompt settings used across different conditions and models. While the core task remains the same, we modified the output conditioning to reduce any failures on specific models.

\begin{itemize}[leftmargin=*]
\item \textbf{Role Specification: } You are a behavior analyst assessing doctor patient interaction to help doctors with communication feedback. You will be given a small part of a transcript of a conversation between a doctor and a patient.

\item \textbf{Job Description: } 
You have to score the \texttt{<signal\_name>} in this text. Look at each behavior and look for deviations from normal or neutral behavior, be it positive (high) or negative (low).

\item \textbf{Scoring Instructions:} [FLAN-T5/Gemma2] Is the \texttt{signal\_name} higher than normal? [Llama] The expected scale tags each behavior with an integer: 0 means below average or not existent, 1 means above average or existent. 
\newline This is updated to "Did you see any presence of \texttt{signal\_name} in this slice?" for Type-II signals.

\item \textbf{Output Format (answer): } [Llama] Return the integer only. [FLAN-T5/Gemma2] Respond only with one word, "yes" or "no".

\item \textbf{Output Format (answer + reasoning):} [Llama] Return the integer first, followed by an explanation of the score in two sentences. [FLAN-T5/Gemma2] Is the \texttt{signal\_name} higher than normal? Start your answer with either Yes or No. Explain why you think it was higher or lower. DO NOT repeat any sentence!

\item \textbf{Few-Shot Examples: }
\begin{itemize}
    \item \textit{Llama:} Here are examples (please interpret speaker turns accurately based on the context, even if the diarization may not be perfect): High provider dominance/assertiveness example: xxx; Low provider dominance/assertiveness example: xxx.
    \item \textit{FLAN-T5/Gemma2:} \#TRANSCRIPT: \{example\}; \#LABEL: \{yes\}; ... \#TRANSCRIPT: \{example\}; \#LABEL: \{no\}; \#TRANSCRIPT: \{example\}; \#LABEL:
\end{itemize}
\end{itemize}

\newpage
\section{Robustness Investigation}
\begin{figure*}[h]
    \centering
    \includegraphics[width=\linewidth]{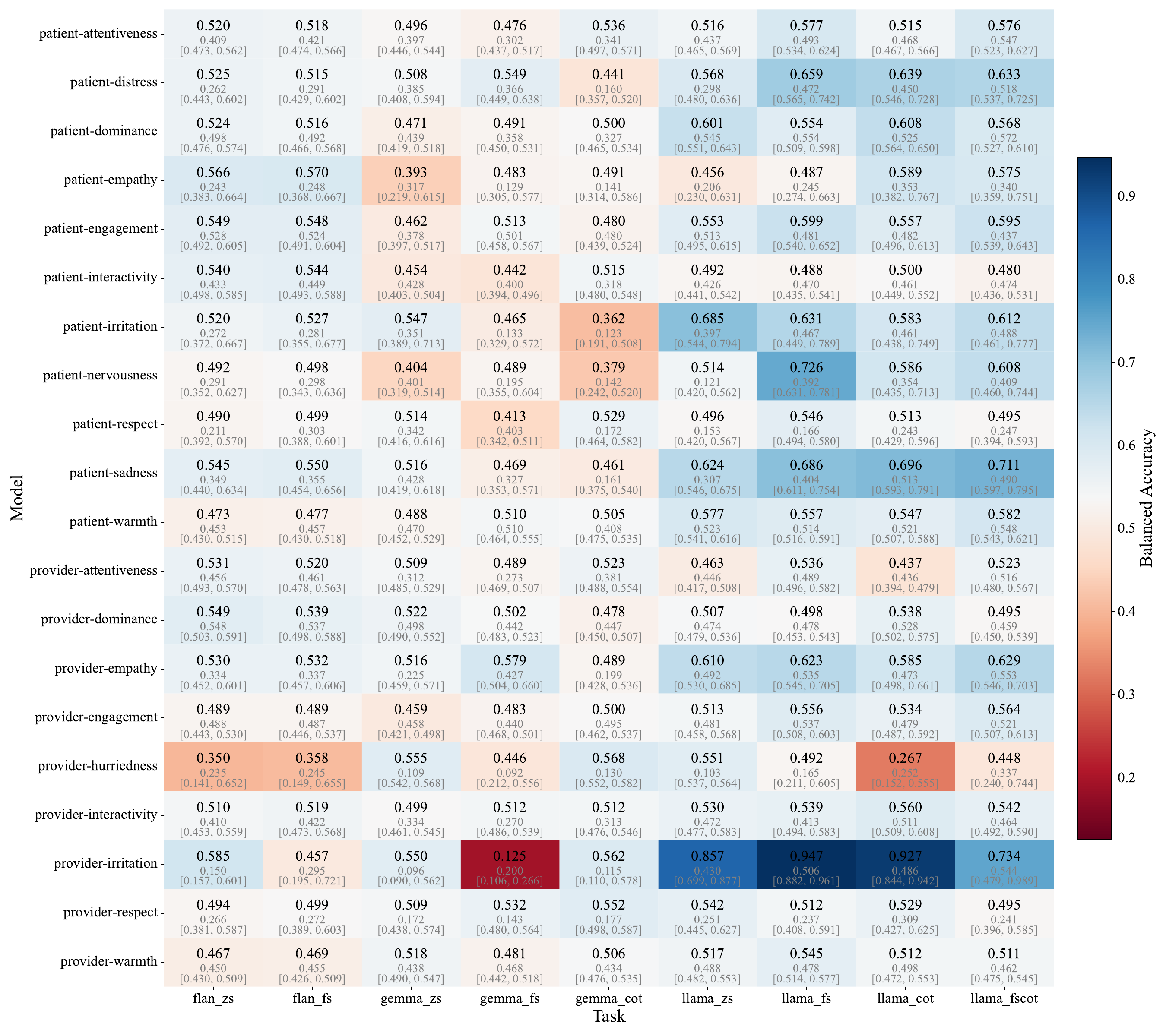}
    \vspace{-2em}
    \caption{Bootstrapped performance for all configuration–task pairs is visualized in this heatmap, with each cell showing mean balanced accuracy, macro-F1, and 95\% confidence intervals (based on balanced accuracy). At least one model per task reliably exceeds chance (0.5), and for most tasks, even the lower confidence bound is above 0.5, highlighting consistent predictive ability. Macro-F1 is provided for completeness. These findings demonstrate that LLMs can feasibly detect trends in social signals across tasks.}
    \label{fig:boostrap}
    \vspace{-2em}
\end{figure*}

Since the dataset is small and imbalanced, in addition to the individual results (Table~\ref{tab:overall_acc}), we conduct reliability estimation by \textit{boostrapping}. As prescribed in~\cite{wilcox2010bootstrap}, we used 599 iterations to resample prediction and label pairs from each configuration and task. Figure~\ref{fig:boostrap} shows the results of this analysis.

For all the social signals at least one of the configurations result in an averaged balanced accuracy over all resamples greater than chance (0.5) with over 65.55\% of all configuration-task pairs reporting above chance behavior. We also note that in 15 of 20 tasks, the lower bound of the 95\% confidence interval is also greater than chance. The macro-F1 scores are a stricter metric and tend to be low in many signals. Overall these results further validate the \textbf{feasibility of tracking social signal levels} using LLMs.

\section{Indicators of a difficult slice}
\label{app:difficult}
We also examine whether certain transcript segments—specific moments within a visit—are more difficult for LLMs when predicting social signals. Understanding which slices are consistently harder to interpret can help clarify when model outputs may be less reliable. These insights are useful for designing systems that surface uncertainty when providing feedback or assessments based on conversational data.

\begin{figure}[h]
\vspace{-1em}
\centering
\includegraphics[width=0.5\linewidth]{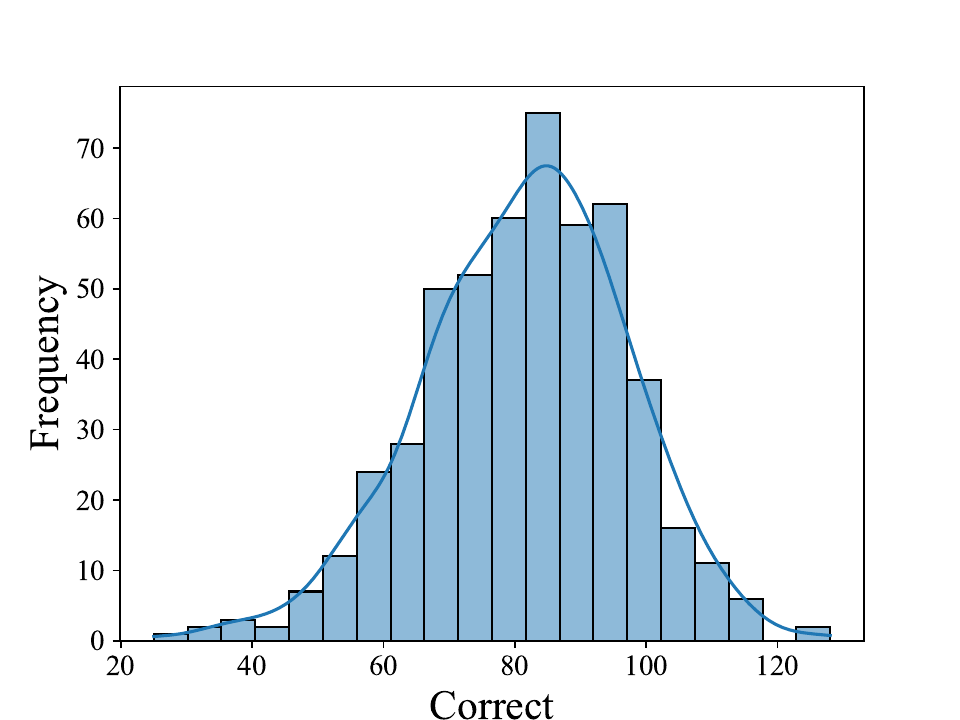}
\caption{Histogram distribution of number of correct predictions per sample. We see that the correct predictions are normally distributed.}
\label{fig:sample_dist}
\vspace{-1em}
\end{figure}

We first aggregate the number of correct predictions across all tasks to the level of individual transcript slices. With 9 model–prompt configurations and 20 tasks, each sample has a maximum of 180 possible predictions. The distribution of correct predictions per sample is approximately normal, as confirmed by a Kolmogorov–Smirnov test (D = 0.048, p = 0.181), with a mean of 81.39 and a standard deviation of 15.52. The number of correct predictions per sample ranges from 25 to 128. Figure~\ref{fig:sample_dist} illustrates this distribution across all transcript slices. To better understand what makes certain samples easier or harder to interpret, we conduct two complementary analyses: one quantitative and one qualitative.

\subsection{Quantitative Differentiation}
Since our analysis is based entirely on textual transcripts, we examine linguistic characteristics to identify differences between high- and low-accuracy samples. To do this, we use LIWC-22~\cite{boyd2022development}, a widely used tool for capturing affective and psychological dimensions of language. LIWC has been validated in related domains such as therapist–patient interactions~\cite{vail2022toward} and support-seeking conversations~\cite{xue2024interaction}.

We define ``easy'' and ``hard'' samples as slices falling in the upper and lower quantiles of the distribution of correct predictions, respectively. We then compare linguistic markers between these two groups using the Mann–Whitney U-test. To account for multiple comparisons across LIWC categories, we apply a Bonferroni correction. Table~\ref{tab:liwc} presents the LIWC categories that significantly differ between “hard” and “easy” slices based on their agreement across all configurations and tasks. We observe that linguistic, cognitive, affective, and social-behavioral descriptors vary meaningfully between these groups. Interestingly, descriptors related to physical and mental health or lifestyle do not show significant differences. This suggests that the difficulty of social signal prediction is influenced less by the clinical content of the conversation and more by general linguistic and affective patterns. This highlights the potential for these findings to generalize across domains beyond healthcare.

\begin{table*}[t]
\centering

\begin{tabular}{lllll}
\toprule
\textbf{Feature} & \textbf{Description (example words)} & \textbf{Hard} & \textbf{Easy} & \textbf{U-statistic} \\
\midrule
Tone & Degree of positive (negative) tone & 43.557 (26.14) & 54.512 (28.86) & 16956.5 * \\
WPS & Words per sentence & 8.378 (4.03) & 11.048 (47.32) & 28704.5 ** \\
function & Function words (the, to, and, I) & 51.483 (9.87) & 47.424 (10.72) & 28676.5 ** \\
you & Second-person pronouns (you) & 3.994 (2.10) & 6.684 (7.31) & 17619.0 * \\
ipron & Impersonal pronouns (that, it, this, what) & 6.845 (2.60) & 5.718 (2.85) & 28618.5 ** \\
det & Determiners (the, at, that, my) &  11.592 (2.91) & 9.974 (4.06) & 28189.0 * \\
article & Articles (a, an, the, alot) & 4.335 (1.67) & 3.624 (1.84) & 28048.0 * \\
conj & Conjunctions (and, but, so, as) &  6.208 (2.33) & 5.231 (2.65) & 28062.0 * \\
Cognition & Cognition (is, was, but, are) &  12.167 (4.07) & 10.377 (4.77) & 27898.0 * \\
cogproc & Cognitive Processes (but, not, if, or, know) &  10.546 (4.03) & 8.747 (4.52) & 28290.0 ** \\
cause & Causation (how, becuase, make, why) &  1.280 (0.88) & 0.937 (0.73) & 28615.5 ** \\
tentat & Tentative (if, or, any, something) &  2.843 (1.52) & 2.205 (1.61) & 28484.5 ** \\
tone\_pos & Positive Tone (good, well, new, love) &  2.667 (3.65) & 5.316 (7.87) & 17581.0 * \\
polite & Politeness (thank(s), please, good morning) &  0.607 (1.79) & 3.545 (8.12) & 16586.0 *** \\
AllPunc & Number of punctuation marks &  32.920 (10.41) & 35.459 (8.77) & 17523.0 * \\
Period & Number of periods  &  13.686 (7.80) & 15.979 (8.01) & 17813.5 * \\
\bottomrule
\end{tabular}

\vspace{1em}
\caption{Table shows the significantly different textual descriptors in the lower quantile (Hard) and higher quantile (Easy) of slices by their overall probability of correctness achieved by our system. The descriptions are the definitions and exemplar words from the LIWC-22 guide~\cite{boyd2022development}. We report the mean (std) for the LIWC feature values for the two groups as well as the Mann-Whitney U statistics with Bonferroni correction. We use the convention: * (p~\textless0.05), ** (p~\textless0.01) and *** (p~\textless0.001)}
\label{tab:liwc}
\vspace{-2em}
\end{table*}

Among linguistic features, more difficult slices tend to contain shorter sentences and exhibit higher usage of function words, impersonal pronouns, determiners, articles, and conjunctions, alongside lower frequencies of second-person pronouns. In the cognitive domain, difficult samples show increased use of words associated with general cognitive processes (e.g., thinking, reasoning), causation, and tentativeness. In terms of affect, difficult samples are characterized by lower overall tone and reduced use of positive emotion words.

While punctuation-related features also differed between groups, we treat these results with caution, as they are likely influenced by transcription artifacts and may not generalize across datasets. One notable finding is that a lower presence of affective and tone-related words significantly reduces the LLM’s ability to detect social signals. This supports the interpretation that LLMs rely, at least in part, on affective linguistic cues when making predictions. The strong role of both cognitive and emotional language in distinguishing sample difficulty further suggests that LLMs are capable of abstracting away from the clinical topic itself and instead focus on general patterns of social behavior. Future work should explore how these trends differ by speaker role (e.g., patient vs. provider) to gain a more nuanced understanding of where and how LLMs succeed or fail in social signal processing.

\subsection{Qualitative Understanding of difficult samples}
To better understand the nature of moments that are challenging for LLMs to assess, we conducted a qualitative analysis of the lowest 10th percentile of samples (n = 50) by performance (see Figure~\ref{fig:sample_dist}). We explored two dimensions of these transcript slices: the content of the conversation and the style of communication. Using open coding, we inductively identified several recurring themes that characterize difficult-to-predict segments:
\begin{itemize}[leftmargin=*]
\item \textit{Anxious patients, reassuring providers: } Many difficult slices involve patients expressing anxiety—whether recalling experiences related to their health, voicing concerns about inherited conditions, or worrying about side effects or addiction to prescribed medication. In these cases, providers often adopt a reassuring role, attempting to reduce the patient’s distress by offering empathetic or explanatory responses. The subtlety and complexity of these emotional dynamics may make them harder for LLMs to interpret accurately.
\item \textit{Patient frustration with themselves or the healthcare System: } Another common theme is patient frustration, often stemming from feeling unheard or misunderstood by previous providers. For example, one patient recounted how multiple specialists attributed all symptoms to “arthritis,” while new symptoms went unacknowledged. In other cases, patients rapidly switched topics mid-sentence, disrupting the flow of the conversation and creating confusion. These disjointed or emotionally charged exchanges may present challenges for models trying to infer coherent social signals.
\item \textit{Actions in parallel to voice: } Several difficult slices involved segments in which providers appeared to be performing physical examinations or procedural tasks. These interactions consisted primarily of brief, instruction-oriented language with minimal dialogue or back-and-forth exchange. Such slices underscore a core limitation of using text-only inputs: without access to nonverbal context or action cues, LLMs struggle to interpret the social dynamics of these moments.
\end{itemize}

These themes reveal specific patterns in patient-provider communication that are difficult for LLMs to track using textual data alone. Recognizing these types of interactions may help inform the development of more adaptive systems—for example, by routing such segments through alternative multi-prompt strategies or incorporating multimodal signals to enhance robustness. Ultimately, this qualitative insight can guide future improvements in LLM-based social signal processing systems for healthcare and beyond.

\section{Model Configurations' impact on LLM performance}
\label{sec:res_config}

Our overall pipeline demonstrates the potential of LLMs to model social behavior. Here we investigate whether different model configurations yield different performance outcomes. This analysis helps us understand the relative impact of design choices and can guide future researchers in selecting optimal configurations for their specific applications.
\newline
\newline
\noindent
\textbf{Approach -- } 
We analyze the effects of four categorical variables: \texttt{model}, \texttt{prompt}, \texttt{configuration}, and \texttt{task}. Here, \texttt{task} refers to each of the 20 social signal prediction tasks. Also here the three models used are FLAN-T5, Gemma2-2B, and LLaMA3.1-405B, and the prompting strategies include Zero-Shot (ZS), Few-Shot (FS), Chain-of-Thought (CoT), and Few-Shot with Chain-of-Thought (FS-CoT). The \texttt{configurations} represent valid model–prompt pairings used during evaluation. To assess the relative contribution of each variable, we analyze the likelihood of a correct prediction (\texttt{correct}) as a function of these settings. A prediction is labeled \texttt{correct} if the model output matches the ground truth label for a given transcript-task pair. We then model the probability of correctness using Generalized Linear Mixed Models (GLMMs) with a binomial distribution. This method allows us to examine main effects while controlling for variability introduced by other factors. This novel framing of understanding alignment allows us to evaluate various factors impacting Human-AI alignment using known potential covariates.
We use the \texttt{BinomialBayesMixedGLM} module from the Statsmodels library.\footnote{Statsmodels documentation: \href{https://www.statsmodels.org/stable/generated/statsmodels.genmod.bayes_mixed_glm.BinomialBayesMixedGLM.html}{BinomialBayesMixedGLM}} Equations~\ref{eq:model}–\ref{eq:config} show the model specifications for each variable under study, with the variable of interest treated as a fixed effect and all others treated as random effects.

This approach provides a more robust statistical framework than simply comparing accuracy metrics, which are often confounded by dataset imbalance or task difficulty. Prior work on LLMs in behavioral sensing~\cite{xu2024mental} has typically relied on such raw accuracy comparisons. By contrast, our analysis accounts for multi-dimensional variability and therefore helps more reliable inferences about the relative effectiveness of different model configurations.





\begin{subequations}
\vspace{-1em}
\begin{gather}
\begin{aligned}
\text{correct} \sim\;& C(\text{model}, \text{ref}=\text{``FLAN''}) - 1 \\
&+ \text{Random}(1+\text{visit}, 1+\text{prompt}, 1+\text{task})
\end{aligned}
\label{eq:model}
\\
\begin{aligned}
\text{correct} \sim\;& C(\text{prompt}, \text{ref}=\text{``ZS''}) - 1 \\
&+ \text{Random}(1+\text{visit}, 1+\text{task}, 1+\text{model})
\end{aligned}
\label{eq:prompt}
\\
\begin{aligned}
\text{correct} \sim\;& C(\text{configuration}, \text{ref}=\text{``FLAN-ZS''}) - 1 \\
&+ \text{Random}(1+\text{visit}, 1+\text{task})
\end{aligned}
\label{eq:config}
\end{gather}
\end{subequations}

\noindent
\textbf{Effects of Model Settings ---}
Table~\ref{tab:config_lmm} presents the comparative effects of different model settings, including model types, prompting strategies, and their combined configurations. We use the smallest model (FLAN-T5) and the simplest prompting strategy (Zero-Shot) as reference baselines to assess the impact of varying each setting. Below, we summarize the key findings from each of these analyses.

\begin{table}[h]
\centering

\small{\begin{tabular}{lrc}
\toprule
 & \textbf{Coef. (log)} & \textbf{OR}\\
\midrule
\multicolumn{3}{l}{\textit{Model (from Eq~\ref{eq:model})}} \\
FLAN-T5 (reference) & -0.360 & [0.698]\\
Gemma2-2b & -0.185 & 0.831\\
LLaMA3.1-405B & 0.345 & 1.412\\
\\
\multicolumn{3}{l}{\textit{Random Var (visit = 0.05, prompt = 0.22, task=0.25)}} \\ 
\midrule
\multicolumn{3}{l}{\textit{Prompting Style (from Eq~\ref{eq:prompt})}} \\
Zero Shot (reference) & -0.356 & [0.700]\\
Few Shot & 0.094 & 1.098\\
Chain of Thought & -0.014 & 0.986\\
Few Shot with Chain of Thought & 0.359 & 1.431\\
\\
\multicolumn{3}{l}{\textit{Random Var (visit = 0.05, task=0.23, model = 0.12)}} \\ 
\midrule
\textit{Configuration (from Eq~\ref{eq:config})}\\
FLAN-ZS (reference) & -0.199 & [0.820]\\
FLAN-FS & -0.310 & 0.733\\
Gemma-ZS & -0.249 & 0.780\\
Gemma-FS & -0.401 & 0.670\\
Gemma-COT & -0.746 & 0.474\\
LLaMA-ZS & -0.178 & 0.837\\
LLaMA-FS & 0.136 & 1.146\\
LLaMA-COT & 0.252 & 1.286\\
LLaMA-FSCOT & 0.402 & 1.495\\
\\
\multicolumn{3}{l}{\textit{Random Var (visit = 0.05, task=0.23)}}\\
\bottomrule
\end{tabular}}
\vspace{1em}
\caption{Table shows the log-odds and Odds Ratios for Binomial Generalized Linear Mixed Models to predict the odds of being correct based on model-type, prompt-type and combinations of the two. The smallest model (FLAN-T5) and simplest prompt (ZS) are used as references. Odds ratios greater than 1 imply increase odds and vice versa.}
\label{tab:config_lmm}
\vspace{-2em}
\end{table}


\begin{enumerate}[leftmargin=*]
\item \textit{LLaMA is 1.4× more likely to be correct, while Gemma underperforms relative to FLAN-T5.} From Equation~\ref{eq:model} in Table~\ref{tab:config_lmm}, we observe that, using FLAN-T5 as the reference, the odds ratio for Gemma2 is 0.831, whereas LLaMA achieves an odds ratio of 1.412. Random effects across visits are low, while moderate variability is observed across prompt styles and tasks. These results indicate that LLaMA, across all prompting styles, consistently demonstrates better understanding of social signals in clinical conversations. In contrast, Gemma2 underperforms FLAN-T5 by approximately 17\%, despite being a larger model. This suggests that FLAN-T5 may have a stronger inductive bias toward affective content (especially Type-I signals), but LLaMA is more reliable overall.
\vspace{1em}

\item \textit{Prompt-augmentation results are comparable to ZS.} According to Equation~\ref{eq:prompt} in Table~\ref{tab:config_lmm}, FS-CoT achieves an odds ratio of 1.431, significantly outperforming the Zero-Shot (ZS) baseline. Few-Shot (FS) prompting shows a modest improvement (1.098), while Chain-of-Thought (CoT) slightly underperforms ZS (0.986). These results suggest that combining examples with reasoning (FS-CoT) yields the greatest benefit in social affective sensing tasks. However, this configuration requires models capable of producing coherent free-text justifications—something only LLaMA could achieve reliably in our setup. Randomization across models and visits has low impact, while variability across tasks is moderate. Together, these findings reinforce the value of combining in-context examples with reasoning to improve performance.
\vspace{1em}

\item \textit{Only LLaMA configurations with prompt augmentations outperform the FLAN-ZS baseline.} Using FLAN-ZS as the reference configuration, Equation~\ref{eq:config} reveals that Gemma-based configurations consistently underperform, with odds ratios of 0.780 (Gemma-ZS), 0.670 (Gemma-FS), and 0.474 (Gemma-CoT). FLAN-FS also shows lower odds (0.733), while LLaMA configurations steadily improve with each augmentation: 0.837 (LLaMA-ZS), 1.146 (LLaMA-FS), 1.286 (LLaMA-CoT), and 1.496 (LLaMA-FSCoT). This configuration-level analysis—where each model–prompt pair is treated as a distinct unit—offers a more balanced comparison, as sample sizes are uniform across conditions. These results underscore that while FLAN-T5 outperforms several larger configurations (including Gemma and LLaMA-ZS), LLaMA benefits most from prompt augmentations and consistently achieves the highest performance when both few-shot examples and reasoning are included.

\end{enumerate}

\newpage
\section{Variation over Race}
\label{app:race}
Table~\ref{tab:fair} summarizes group-level differences in social signal distributions and model performance across visits involving White and non-White patients. The first set of columns reports the proportion of slices labeled as high for each signal, while the “Statistical Difference” column quantifies their relative divergence. Provider warmth and engagement exhibit the most notable distributional disparities, suggesting systematic variation in perceived affective behaviors across patient groups. The rightmost columns report balanced accuracies averaged over all configurations, illustrating that while performance is generally comparable across groups, some signals—such as provider empathy, patient empathy, and distress—show group-specific sensitivity. Bolded values ($>$0.55) indicate reliable above-chance tracking, whereas underlined values reflect cases where LLM-based inference underperforms a majority baseline.

\begin{table*}[h]
\begin{tabular}{lcc >{\centering\arraybackslash}p{2.1cm} >{\centering\arraybackslash}p{2.1cm} >{\centering\arraybackslash}p{2.1cm}}
\toprule
\textbf{Social Signal} & \multicolumn{2}{c}{\textbf{Labels}} & \textbf{Statistical} & \multicolumn{2}{c}{\textbf{Balanced Accuracy}} \\
 & \textbf{White} & \textbf{Non-White} & \textbf{Difference} & \textbf{White} & \textbf{Non-White} \\
\midrule
Provider Dominance & 0.666 (0.47) & 0.761 (0.43) & 1.597  & 0.505 (0.03) & \bf 0.557 (0.07) \\
Provider Attentiveness & 0.332 (0.47) & 0.326 (0.47) & 0.975  & 0.501 (0.03) & 0.514 (0.07) \\
Provider Warmth * & 0.526 (0.50) & 0.337 (0.48) & 0.457 & 0.508 (0.03) & \underline{0.481 (0.05)} \\
Provider Engagement ** & 0.808 (0.39) & 0.630 (0.49) & 0.406 & 0.509 (0.03) & 0.513 (0.06) \\
Provider Empathy & 0.079 (0.27) & 0.087 (0.28) & 1.105  & \bf 0.584 (0.07) & \underline{0.494 (0.06)} \\
Provider Respect & 0.048 (0.21) & 0.033 (0.18) & 0.667  & 0.512 (0.03) & \bf 0.560 (0.14) \\
Provider Interactivity & 0.212 (0.41) & 0.293 (0.46) & 1.548  & 0.531 (0.02) & 0.502 (0.03) \\
Patient Dominance & 0.293 (0.46) & 0.152 (0.36) & 0.433  & 0.530 (0.04) & \bf 0.571 (0.10) \\
Patient Attentiveness & 0.267 (0.44) & 0.130 (0.34) & 0.412  & 0.528 (0.03) & 0.500 (0.06) \\
Patient Warmth & 0.454 (0.50) & 0.293 (0.46) & 0.499  & 0.523 (0.04) & 0.521 (0.05) \\
Patient Engagement & 0.846 (0.36) & 0.859 (0.35) & 1.105  & 0.547 (0.05) & 0.505 (0.08) \\
Patient Empathy & 0.012 (0.11) & 0.011 (0.10) & 0.903  & \underline{0.494 (0.05)} & \bf 0.604 (0.17) \\
Patient Respect & 0.048 (0.21) & 0.022 (0.15) & 0.440  & 0.502 (0.04) & \underline{0.473 (0.14)} \\
Patient Interactivity & 0.219 (0.41) & 0.293 (0.46) & 1.484  & 0.502 (0.04) & 0.476 (0.05) \\
Provider Hurriedness & 0.012 (0.11) & 0.000 (0.00) & 0.000  & \underline{0.450 (0.10)} & \underline{0.215 (0.15)} \\
Provider Irritation & 0.005 (0.07) & 0.000 (0.00) & 0.000  & \bf 0.637 (0.26) & 0.507 (0.36) \\
Patient Irritation & 0.022 (0.15) & 0.011 (0.10) & 0.497  & 0.542 (0.11) & \bf 0.589 (0.15) \\
Patient Nervousness & 0.022 (0.15) & 0.043 (0.21) & 2.056  & 0.526 (0.16) & 0.511 (0.09) \\
Patient Sadness & 0.046 (0.21) & 0.043 (0.21) & 0.950  & \bf 0.588 (0.11) & \bf 0.566 (0.10) \\
Patient Distress & 0.043 (0.20) & 0.109 (0.31) & 2.696  & \bf 0.584 (0.11) & 0.515 (0.05) \\
\bottomrule
\end{tabular}
\vspace{1em}
\caption{Table shows the differences in social signals across White and non-White patients. We see that when binarized, the social signal labels of provider warmth and engagement are significantly different. We also show the averaged balanced accuracies across all configurations on the two subset of visits reported as mean (std); \textbf{bolded values} show the best configurations ($>$0.55). \underline{Underlined Values} are settings where LLM performance to track the social signal levels was worse than majority classifier (0.5)}
\label{tab:fair}
\vspace{-1em}
\end{table*}

\newpage
\section{Variation over Time within Visit}
\label{appendix:time-of-slice}
We divide each visit into three segments—start, middle, and end. The first and last slice of a visit are labeled as start and end, while all intermediate slices constitute the middle. We compare the distribution of human annotations and model performance as shown in Table~\ref{tab:seg_combined}.

\begin{table*}[h]
\centering
\resizebox{\textwidth}{!}{
\begin{tabular}{lccclcccc}
\toprule
\textbf{Social Signal} & \textbf{Start (Dist)} & \textbf{Mid (Dist)} & \textbf{End (Dist)} & \textbf{Chi-square $\chi^2(2)$} & \textbf{Start (Acc)} & \textbf{Mid (Acc)} & \textbf{End (Acc)} \\
\midrule
\textbf{Provider Dominance} & 0.527 (0.50) & 0.752 (0.43) & 0.582 (0.50) & \bar{21.520} *** & \underline{0.493 (0.03)} & 0.514 (0.03) & 0.510 (0.03) \\
Provider Attentiveness & 0.363 (0.48) & 0.324 (0.47) & 0.330 (0.47) & \bar{0.477} & 0.507 (0.04) & \underline{0.498 (0.03)} & 0.523 (0.07) \\
\textbf{Provider Warmth} & 0.714 (0.45) & 0.418 (0.49) & 0.549 (0.50) & \bar{26.374} *** & 0.514 (0.06) & 0.508 (0.03) & 0.500 (0.02) \\
Provider Engagement & 0.769 (0.42) & 0.791 (0.41) & 0.714 (0.45) & \bar{2.401} & \underline{0.495 (0.05)} & \underline{0.497 (0.03)} & 0.530 (0.08) \\
Provider Empathy & 0.033 (0.18) & 0.100 (0.30) & 0.055 (0.23) & \bar{5.300} & \underline{0.487 (0.12)} & 0.566 (0.07) & 0.542 (0.05) \\
Provider Respect & 0.099 (0.30) & 0.030 (0.17) & 0.055 (0.23) & \bar{7.674} & \underline{0.491 (0.06)} & 0.540 (0.03) & 0.574 (0.08) \\
\textbf{Provider Interactivity} & 0.055 (0.23) & 0.303 (0.46) & 0.110 (0.31) & \bar{33.570} *** & \underline{0.474 (0.09)} & 0.515 (0.02) & 0.506 (0.04) \\
Patient Dominance & 0.253 (0.44) & 0.276 (0.45) & 0.275 (0.45) & \bar{0.197} & 0.542 (0.05) & 0.544 (0.06) & 0.507 (0.04) \\
Patient Attentiveness & 0.209 (0.41) & 0.245 (0.43) & 0.253 (0.44) & \bar{0.620} & 0.544 (0.04) & 0.529 (0.04) & 0.503 (0.05) \\
Patient Warmth & 0.495 (0.50) & 0.394 (0.49) & 0.484 (0.50) & \bar{4.354} & \underline{0.499 (0.07)} & 0.540 (0.05) & 0.500 (0.06) \\
Patient Engagement & 0.813 (0.39) & 0.882 (0.32) & 0.747 (0.44) & \bar{10.795} & \underline{0.478 (0.04)} & 0.556 (0.07) & \underline{0.490 (0.07)} \\
Patient Empathy & 0.000 (0.00) & 0.015 (0.12) & 0.011 (0.10) & -- & \underline{0.368 (0.19)} & 0.532 (0.06) & \underline{0.367 (0.24)} \\
\textbf{Patient Respect} & 0.033 (0.18) & 0.012 (0.11) & 0.176 (0.38) & \bar{44.923} *** & 0.569 (0.10) & 0.484 (0.09) & 0.517 (0.06) \\
\textbf{Patient Interactivity} & 0.088 (0.28) & 0.303 (0.46) & 0.110 (0.31) & \bar{27.684} *** & 0.521 (0.07) & \underline{0.485 (0.03)} & 0.467 (0.08) \\
Provider Hurriedness & 0.011 (0.10) & 0.009 (0.10) & 0.011 (0.10) & \bar{0.044} & \underline{0.407 (0.25)} & \underline{0.486 (0.13)} & \underline{0.377 (0.21)} \\
Provider Irritation & 0.000 (0.00) & 0.006 (0.08) & 0.000 (0.00) & -- & \underline{0.466 (0.37)} & 0.645 (0.26) & \underline{0.477 (0.32)} \\
Patient Irritation & 0.033 (0.18) & 0.018 (0.13) & 0.022 (0.15) & \bar{0.743} & 0.539 (0.11) & 0.542 (0.11) & 0.529 (0.16) \\
Patient Nervousness & 0.033 (0.18) & 0.027 (0.16) & 0.011 (0.10) & \bar{1.021} & \underline{0.471 (0.19)} & 0.523 (0.09) & 0.570 (0.32) \\
Patient Sadness & 0.055 (0.23) & 0.045 (0.21) & 0.033 (0.18) & \bar{0.518} & 0.547 (0.19) & 0.589 (0.07) & 0.593 (0.16) \\
Patient Distress & 0.066 (0.25) & 0.058 (0.23) & 0.033 (0.18) & \bar{1.106} & 0.571 (0.09) & 0.550 (0.07) & 0.551 (0.09) \\
\midrule
\textbf{Average (Acc)} & -- & -- & -- & -- & 0.499 (0.05) & 0.532 (0.04) & 0.507 (0.06) \\
\bottomrule
\end{tabular}}
\vspace{1em}
\caption{Merged summary of manually-coded binary label distributions and model accuracy across visit regions (start, middle, end). Distributions are shown as mean (SD). Significant Chi-squared tests with Bonferroni correction are marked (*** $p<0.001$). Accuracy scores \underline{underlined} fall below the 0.5 balanced accuracy threshold; bolded signals show significant distributional differences.}
\label{tab:seg_combined}
\end{table*}

Table~\ref{tab:seg_combined} shows the distribution of social behaviors across these segments and measures the extent to which they differ using a Chi-squared test with Bonferroni correction. Significant effects ($p<0.001$) are observed for provider dominance, warmth, and interactivity, as well as for patients’ respect and interactivity, suggesting that interactional dynamics evolve notably across the course of the visit. The right-hand columns report the average balanced accuracy (and standard deviation) of model predictions within each segment. As shown, model performance remains consistent overall, with a slight improvement during the middle of the visit—potentially reflecting greater conversational richness or behavioral stability at that stage. Underlined values denote cases where balanced accuracy fails to exceed the 0.5 threshold.

\newpage
\section{Ensembling Experiments}
\label{appendix:ensembling}
Table~\ref{tab:highest_scores_combined} summarizes the best-performing configurations for each social signal across three evaluation modes: (1) \textit{Raw} or best performing individual model; (2) \textit{LOGO}, a leave-one-group-out logistic regression baseline without weighting; and (3) \textit{Ensemble}, which aggregates predictions using agreement-weighted fusion of models with optimized classifier and selection strategies. Results are reported as balanced accuracy with standard deviation across held-out groups. As shown, ensemble methods consistently outperform both raw and LOGO baselines across most social signals demonstrating the benefit of leveraging complementary strengths of multiple prompting configurations and model families. Signals such as provider dominance, interactivity, and empathy—as well as patient distress and respect—show substantial gains under ensemble aggregation, highlighting their stability across model architectures and prompt variations.

\begin{table*}[h]
\centering
\resizebox{\textwidth}{!}{
\small
\begin{tabular}{lcccclcccc}
\toprule
\textbf{Signal} & \textbf{Category} & \textbf{Method} & \textbf{Score} & \textbf{Std} &
\textbf{Signal} & \textbf{Category} & \textbf{Method} & \textbf{Score} & \textbf{Std} \\
\midrule
\multirow{3}{*}{provider-dominance} & Raw & flan\_zs & 0.549 & 0.024 &
\multirow{3}{*}{provider-hurriedness} & Raw & gemma\_cot & 0.568 & 0.105 \\
 & LOGO & LOGO & 0.603 & 0.130 &
 & LOGO & LOGO & 0.664 & 0.240 \\
 & Ensemble & pr\_wt\_dtc\_g & \textbf{0.609} & 0.023 &
 & Ensemble & pr\_wt\_lr\_l3p1 & \textbf{0.669} & 0.075 \\
\midrule
\multirow{3}{*}{provider-attentiveness} & Raw & llama\_fs & 0.536 & 0.034 &
\multirow{3}{*}{provider-irritation} & Raw & llama\_fs & \textbf{0.947} & 0.262 \\
 & LOGO & LOGO & \textbf{0.624} & 0.090 &
 & LOGO & LOGO & 0.901 & 0.180 \\
 & Ensemble & pr\_mac\_lr\_l3p1 & 0.553 & 0.022 &
 & Ensemble & pr\_wt\_lr\_all & 0.900 & 0.131 \\
\midrule
\multirow{3}{*}{provider-warmth} & Raw & llama\_fs & 0.545 & 0.026 &
\multirow{3}{*}{patient-irritation} & Raw & llama\_zs & 0.685 & 0.096 \\
 & LOGO & LOGO & 0.558 & 0.100 &
 & LOGO & LOGO & 0.659 & 0.190 \\
 & Ensemble & pr\_wt\_dtc\_l3p1 & \textbf{0.626} & 0.036 &
 & Ensemble & pr\_wt\_dtc\_g & \textbf{0.721} & 0.111 \\
\midrule
\multirow{3}{*}{provider-engagement} & Raw & llama\_fscot & 0.564 & 0.035 &
\multirow{3}{*}{patient-nervousness} & Raw & llama\_fs & 0.726 & 0.106 \\
 & LOGO & LOGO & 0.609 & 0.190 &
 & LOGO & LOGO & 0.650 & 0.230 \\
 & Ensemble & pr\_wt\_dtc\_g & \textbf{0.632} & 0.037 &
 & Ensemble & pr\_wt\_dtc\_all & \textbf{0.733} & 0.110 \\
\midrule
\multirow{3}{*}{provider-empathy} & Raw & llama\_fscot & 0.629 & 0.051 &
\multirow{3}{*}{patient-sadness} & Raw & llama\_fscot & \textbf{0.711} & 0.098 \\
 & LOGO & LOGO & 0.628 & 0.130 &
 & LOGO & LOGO & 0.681 & 0.150 \\
 & Ensemble & pr\_wt\_dtc\_top3 & \textbf{0.676} & 0.067 &
 & Ensemble & pr\_wt\_lr\_top2 & 0.703 & 0.109 \\
\midrule
\multirow{3}{*}{provider-respect} & Raw & gemma\_cot & 0.552 & 0.021 &
\multirow{3}{*}{patient-distress} & Raw & llama\_fs & 0.659 & 0.072 \\
 & LOGO & LOGO & 0.488 & 0.060 &
 & LOGO & LOGO & 0.644 & 0.100 \\
 & Ensemble & pr\_mac\_lr\_top3 & \textbf{0.635} & 0.049 &
 & Ensemble & pr\_wt\_lr\_top3 & \textbf{0.723} & 0.092 \\
\midrule
\multirow{3}{*}{provider-interactivity} & Raw & llama\_cot & 0.560 & 0.020 &
\multirow{3}{*}{patient-dominance} & Raw & llama\_cot & \textbf{0.608} & 0.049 \\
 & LOGO & LOGO & 0.484 & 0.060 &
 & LOGO & LOGO & 0.587 & 0.080 \\
 & Ensemble & pr\_wt\_dtc\_top2 & \textbf{0.673} & 0.047 &
 & Ensemble & pr\_wt\_lr\_all & 0.607 & 0.038 \\
\midrule
\multirow{3}{*}{patient-attentiveness} & Raw & llama\_fs & 0.577 & 0.033 &
\multirow{3}{*}{patient-warmth} & Raw & llama\_fscot & 0.582 & 0.042 \\
 & LOGO & LOGO & 0.589 & 0.030 &
 & LOGO & LOGO & 0.564 & 0.060 \\
 & Ensemble & pr\_mac\_lr\_top3 & \textbf{0.616} & 0.032 &
 & Ensemble & pr\_wt\_dtc\_top2 & \textbf{0.582} & 0.030 \\
\midrule
\multirow{3}{*}{patient-engagement} & Raw & llama\_fs & 0.599 & 0.047 &
\multirow{3}{*}{patient-empathy} & Raw & llama\_cot & 0.589 & 0.067 \\
 & LOGO & LOGO & 0.554 & 0.040 &
 & LOGO & LOGO & 0.464 & 0.180 \\
 & Ensemble & pr\_wt\_dtc\_top5 & \textbf{0.627} & 0.024 &
 & Ensemble & pr\_mac\_dtc\_top5 & \textbf{0.616} & 0.086 \\
\midrule
\multirow{3}{*}{patient-respect} & Raw & llama\_fs & 0.546 & 0.037 &
\multirow{3}{*}{patient-interactivity} & Raw & flan\_fs & 0.544 & 0.035 \\
 & LOGO & LOGO & 0.575 & 0.100 &
 & LOGO & LOGO & 0.599 & 0.130 \\
 & Ensemble & pr\_wt\_dtc\_top3 & \textbf{0.754} & 0.083 &
 & Ensemble & pr\_mac\_dtc\_all & \textbf{0.626} & 0.047 \\
\bottomrule
\end{tabular}}
\vspace{1em}

\caption{Best-performing configurations for each social signal. For each signal, we report the highest balanced accuracy across three approaches: Raw (best individual model–prompt configuration), LOGO (leave-one-group-out logistic regression without weighting), and Ensemble (agreement-weighted ensemble with best aggregator and model selection). Method abbreviations: \textit{flan/gemma/llama} = model type; \textit{zs/fs/cot/fscot} = prompting strategy (zero-shot, few-shot, chain-of-thought, few-shot with CoT); \textit{pr\_wt/pr\_mac} = weighted precision or macro-averaged precision confidence weighting; \textit{lr/dtc} = logistic regression or decision tree classifier aggregator; \textit{all/g/l3p1/top2/top3/top5} = model selection strategy (all models, Gemma only, LLaMA with all prompts, or top-$k$ models by agreement). Bold indicates the best score for each signal.}
\label{tab:highest_scores_combined}
\end{table*}